\documentclass[pdflatex]{IEEEtran}
\IEEEoverridecommandlockouts
\usepackage{cite}
\usepackage{amsmath,amssymb,amsfonts}
\usepackage[linesnumbered,lined,commentsnumbered]{algorithm2e}
\usepackage[utf8]{inputenc}
\usepackage{setspace}

\usepackage{graphicx}
\usepackage{textcomp}
\usepackage{xcolor}
\DeclareUnicodeCharacter{2009}{FIXME}
\def\BibTeX{{\rm B\kern-.05em{\sc i\kern-.025em b}\kern-.08em
    T\kern-.1667em\lower.7ex\hbox{E}\kern-.125emX}}
\begin{document}

\title{Closed-loop Control of Catalytic Janus Microrobots\\

}

\author{\IEEEauthorblockN{Max Sokolich,
David Rivas, 
Zameer Hussain Shah,
Sambeeta Das\IEEEauthorrefmark{1}}

\IEEEauthorblockA{Department of Mechanical Engineering,
University of Delaware, Newark, DE 19717 USA}

}

\maketitle

\begin{abstract}
We report a closed-loop control system for paramagnetic catalytically self-propelled Janus microrobots. We achieve this control by employing electromagnetic coils that direct the magnetic field in a desired orientation to steer the microrobots. The microrobots move due to the catalytic decomposition of hydrogen peroxide, during which they align themselves to the magnetic torques applied to them. Because the angle between their direction of motion and their magnetic orientation is a priori unknown, an algorithm is used to determine this angular offset and adjust the magnetic field appropriately. The microrobots are located using real-time particle tracking that integrates with a video camera. A target location or desired trajectory can be drawn by the user for the microrobots to follow.
\end{abstract}


\section{Introduction}
Microrobots are tiny machines designed to perform specific tasks in complex environments.\cite{b1} These nanomachines hold the key to the future of effective biomedicine and a clean environment.\cite{b2,b3} Microrobotics is still an infant technology, and more fundamental and applied work is needed before its widespread use.\cite{b4} On the fundamental side, understanding the active motion of microrobots is of prime significance. Microrobots can be actuated by a chemical fuel such as peroxide or by an external field such as a magnetic field or light, etc.\cite{b5} Chemically powered microrobots have been the center of attention since their discovery in 2004.\cite{b6} These micromotors utilize fuel from their surroundings as a source of energy to drive their motion.\cite{b7} Typically, these micron-sized particles are equipped with a catalyst that can decompose the fuel. In the classic design, polymeric microspheres are half-coated with platinum to fabricate asymmetric particles, also known as Janus colloids.\cite{b8} The non-symmetric nature of these particles results in an asymmetric breakdown of fuel molecules around the particle. This non-uniform reaction around the particle results in an unequal pressure on the particle that leads to its directional motion.\cite{b9}

Even though chemically propelled particles follow a trajectory, their direction of motion is not controlled and often approximates a random walk at long times due to thermal rotational fluctuations. To get control over the directionality, it is a common practice to include a magnetic component in a chemically powered micromotor.\cite{b11}This simple strategy allows the chemically powered micromotors to move in a specific direction under the influence of a magnetic field.  %Since the applied field is global in nature, all the active particles receive the same signal; hence controlled motion of individual agents is not achieved.\cite{b12} However, for practical applications such as targeted therapy and micromanipulation, etc. a point-to-point control is highly desired.\cite{b13} Therefore, methods to control the individual magneto-catalytic micromotors are of great significance.

Magnetic fields have been used to propel and steer microrobots for over a decade.\cite{b14} They offer a facile approach for moving microrobots towards desired locations. Palacci demonstrated that a light activated colloidal particle could be steered magnetically to a colloidal particle and then transport it to a targeted location.\cite{b15} Similarly, Baraban and co-workers reported that homogeneous magnetic fields on the order of milli-Teslas could be used to control the direction of motion of Janus micromotors.\cite{b16} Weak magnetic fields have also been shown to control the motion and localization of Pt-SiO2 Janus micromotors in 2D as well as 3D spaces.\cite{b17} In a more advanced model of magnetically controlled navigation, Joseph Wang’s group have developed smart microrobots capable of autonomous navigation in complex environments and traffic scenarios.\cite{b18} %Despite these developments, a precise control over motion of the catalytically powered microrobots is limited.\cite{b19} %control over multiple microrobots is limited? That last paper you cited does seem to show control of catatlytic bots quite convincingly

For practical applications such as targeted therapy and micromanipulation, it is highly desired to precisely control the microrobot motion from one point to another point.\cite{b10} Attaining greater precision than with manual, open-loop, control, can be attained by using an automated control strategy. Automation is also advantageous due to its greater reproducibility, predictability, and ease of use. Therefore, one of the most attractive approaches for precise control over microrobots is the so called closed-loop control also known as feedback control. In closed-loop control the response is continuously compared with the desired output and modified to minimize any deviations.\cite{b20} Closed-loop control has been utilized to control the motion of microrobots of different shapes. \cite{b26} Khalil and co-workers have extensively studied the closed-loop control of magnetically responsive chemically propelled microjets.\cite{b21,b23} The authors controlled orientation of the microjets with external magnetic torques while the linear motion towards a reference was controlled by a thrust generated from the chemically produced oxygen bubbles and a magnetic field gradient. Similarly, Marino et al.\cite{b24} have reported the closed-loop control studies of cylindrical-shaped microbots. The authors presented different methods to control the uncertainties in electromagnetic force generation and drag forces. Recently, Jiang et al.\cite{b25} have developed a closed-loop control system to control the motion of a vortex-like magnetic microswarm. They employed commercial servo amplifiers to improve the performance of their system. Interestingly, most of the work published on the closed-loop control of microrobots is done on particles of shapes that are a specialty of the group. Since the most common form of catalytic microrobots is a spherical Janus colloid, closed-loop control systems for these particles is of particular importance.

In this work, we demonstrate a simple control algorithm for transporting paramagnetic catalytic Janus micromotors to target locations in the workspace. This is done using a custom tracking program written in python, and the ModMag microrobotic manipulation device. The algorithm is also extended towards more complicated trajectories. We demonstrate this by drawing curved paths on the video display which the microrobots follow.

\section{Methods}

The catalytically propelled Janus microrobots were made of commercially available paramagnetic microspheres (Spherotech, Magnetic beads 4.7 µm in diameter). These particles were spread on a plasma-cleaned glass substrate to make a monolayer. Then these particles were half-coated with a 15 nm nickel layer and a 40 nm platinum layer using e-beam deposition. Finally, the Janus particles were collected by mopping the glass substrate with a wet lens cleaning paper. The paper was vortexed in a 1.5 ml vial containing 0.5 ml DI water which resulted in a dilute suspension of the Janus colloids. The particles were used for the closed-loop experiments without any further treatment.

\begin{figure}[ht]
  \centering
  \includegraphics[width=.95\linewidth]{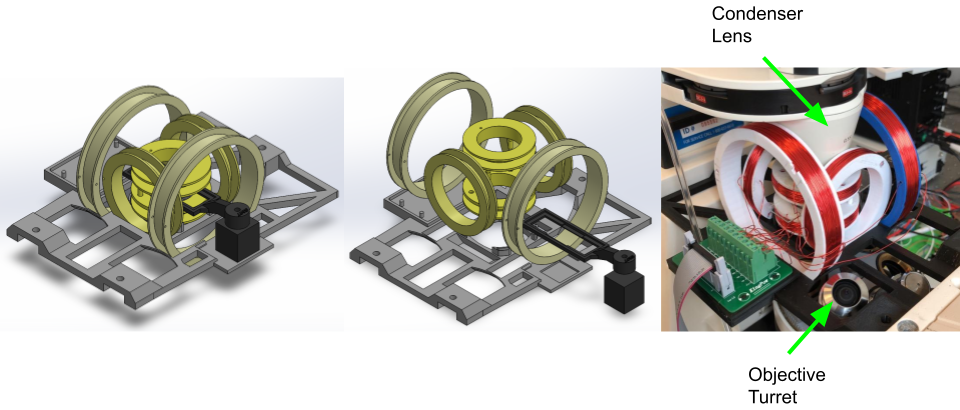}
  \caption {Experimental setup showing a 3 dimensional helmholtz coil configuration electromagnetic mounted on a Zeiss Axiovert 200 inverted microscope. a) CAD diagram, b) CAD exploded view, c) coils mounted on microscope.}
  \label{t}
\end{figure}

To study the motion of Janus microrobots, 50 µl of the particle suspension was dropped onto a plasma-cleaned glass substrate. The particles were allowed to settle down, then predetermined amounts of hydrogen peroxide (Fischer Scientific, 30\%) were added to the particles on the substrate. The particles started to move immediately upon addition of the hydrogen peroxide. The propulsion was the result of non-uniform chemical decomposition of hydrogen peroxide around the particles. The trajectories of the microrobots in the absence of the magnetic field were curved. Controlling the direction of motion of the microrobots was achieved by applying magnetic torques. Uniform magnetic fields on the order of a milli-Tesla are produced with a custom-built  electromagnetic coil system that consists of six independent electromagnetic coils as shown in Figure \ref{t}. Although the coils are capable of applying a magnetic field in any 3D orientation, we only focus on fields in the horizontal plane in this work.

Images are acquired using a camera (details of the camera here) which displays a video feed in Amscope software. Python code was written that acquires a screen shot of the video feed and continuously displays the images as well as highlighting objects that are identified as microrobots in a figure window. The object finding code can be adjusted to account for microrobots of different sizes or intensities.

\begin{figure}
  \centering
  \includegraphics[width=.95\linewidth]{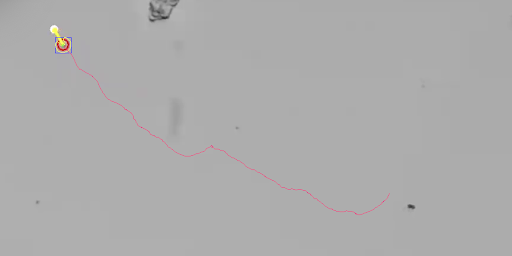}
  \caption {Experimental trajectory of a 4.7 um in diameter micromotor. The micro motor is highlighted in red.  }
  \label{track}
\end{figure}
\section{Results and Discussion}

The paramagnetic Janus microrobots we use generally do not move in the direction of the applied magnetic field due to the angular offset between their magnetic moment vector and the direction of catalytic propulsion. The application of a magnetic field results in the microrobot's trajectory changing from curved to a straight line trajectory. However, the direction of its trajectory is a priori unknown. As a result, an algorithm (Algorithm 1) is used to determine the appropriate magnetic field direction based on the measured directional motion of the microrobot after an initial magnetic field is applied in the x-direction (chosen arbitrarily). The basic logic behind the algorithm is shown in figure 4. 

\begin{figure}[ht]
  \centering
  \includegraphics[width=\linewidth]{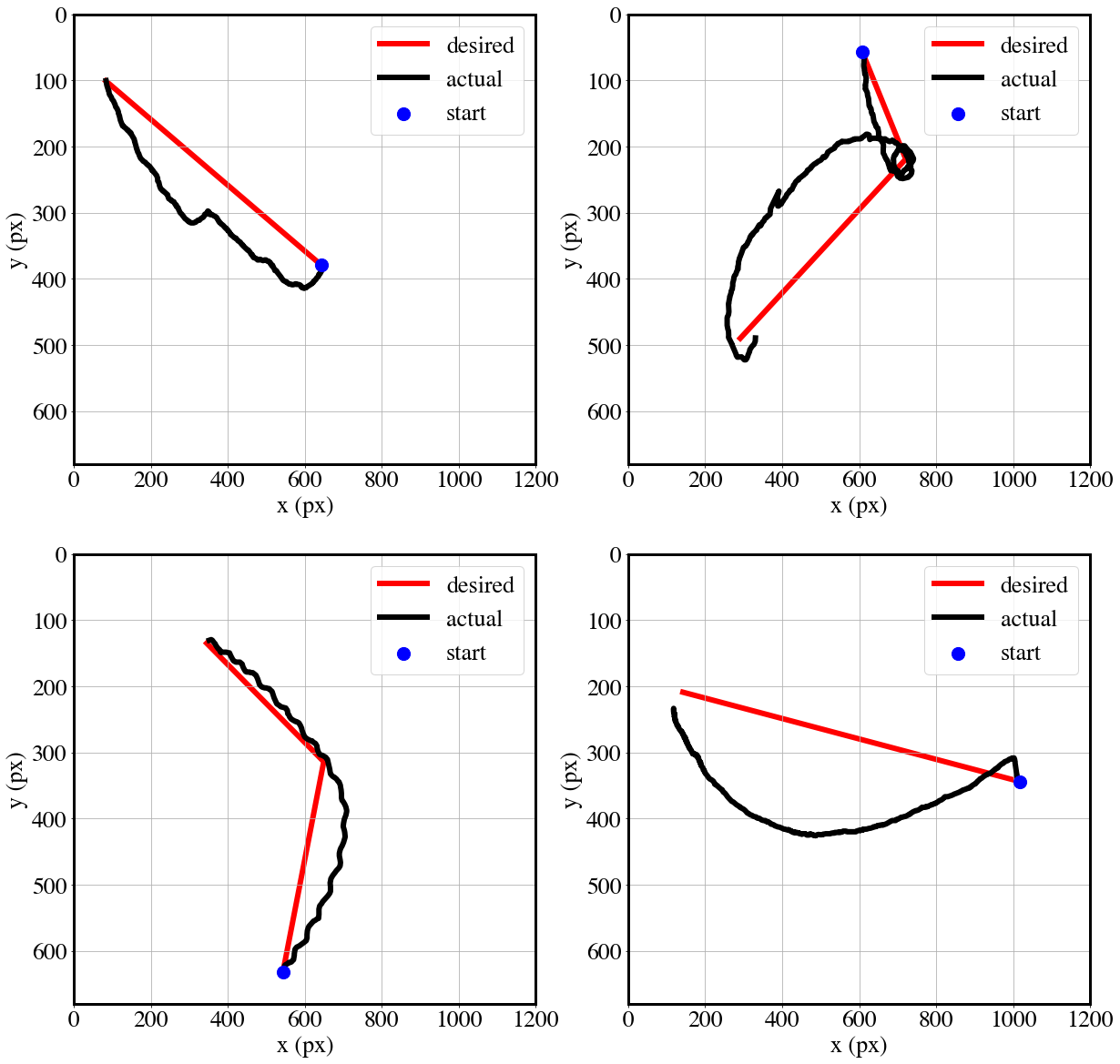}
  \caption {Four experimental trajectories of platinum coated diffusiophoretic Janus micromotors. The blue point indicates the starting position of the micromotor. Algorithm 1 is applied resulting in the black trajectories.}
  \label{traj}
\end{figure}

\begin{figure*}[ht!]
  \begin{center}
  % \centering
    \includegraphics[width=0.9\textwidth]{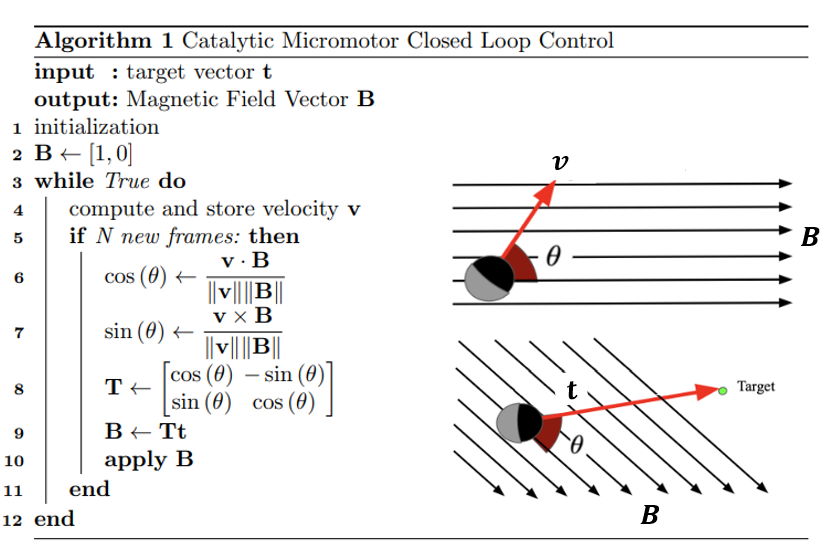}
      \caption{Algorithm 1. Black lines in the visual representation on the right hand side represent applied magnetic field lines. Red lines indicate the micro-motors velocity direction. Once a target location is selected by the user, the code applies a magnetic field that accounts for the angular offset between the field and the direction of microrobot motion.}
    \label{Figure Cell Images}
  \end{center}
\end{figure*}

\begin{figure*}[ht]
  \begin{center}
  % \centering
    \includegraphics[width=0.9\textwidth]{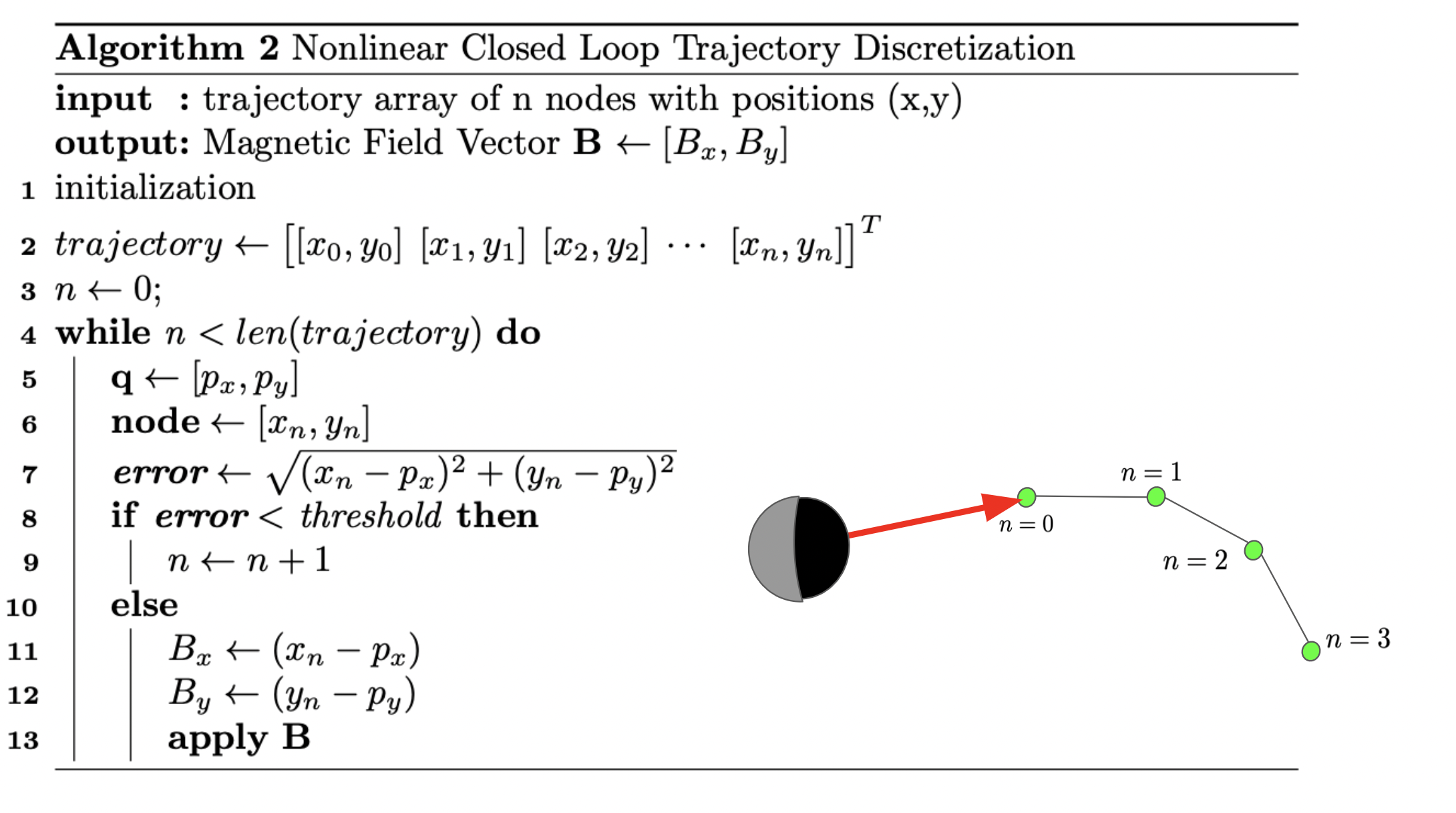}
      \caption{Algorithm 2.  Green points represent that nodes of the trajectory array and the red arrow indicates the micro-motors velocity vector. The microrobot is guided to each node in succession using Algorithm 1.
      }
    \label{Figure }
  \end{center}
\end{figure*}

Algorithm 1 takes inputs of the target vector which points from the microrobot to the target location. The microrobot location is determined using either "Trackpy" or open cv particle locating code in python. The user selects a microrobot by clicking in its proximity with the cursor. Once a microrobot is selected, the code continuously tracks the microrobot by selecting the object that is closest to the previous microrobot's location. To increase the speed of the code, the image is cropped around the center of the microrobot so that the microrobot can be found more rapidly in the following frames. The previous N positions of the microrobot are stored and used to compute an average velocity. The number N essentially acts to reduce the effect of noise when calculating the velocity unit vector. Although currently this value is inputted by the user, in the future this could be determined automatically based upon the estimated signal to noise of the microrobot's velocity. The target location is selected by the user by right-clicking on a location on the video display. Once a target location is selected, a magnetic field is applied in the x-direction. After N positions of the microrobot have been tracked, the algorithm corrects for the difference between the applied magnetic field vector ($\bf{B}$) and the calculated direction of motion of the microrobot, $\bf{v}$. This allows a transformation rotation matrix to be applied to the target vector, $\bf{t}$, which sets the new direction of the magnetic field and accounts for the angular offset between the desired direction of motion of the microrobot and its measured direction of motion. The entries of the transformation matrix, $\cos{(\theta})$ and $\sin{(\theta})$, are determined by taking the dot and cross products, respectively, of the target vector and the magnetic field vector. This updated magnetic field vector is used to apply appropriate current to the coils that will generate the corresponding magnetic field. After another N positions of the microrobot have been found, the magnetic field is updated again based on the latest applied field and velocity and the process continuously repeats.

\begin{figure}[ht]
  \centering
  \includegraphics[width=\linewidth]{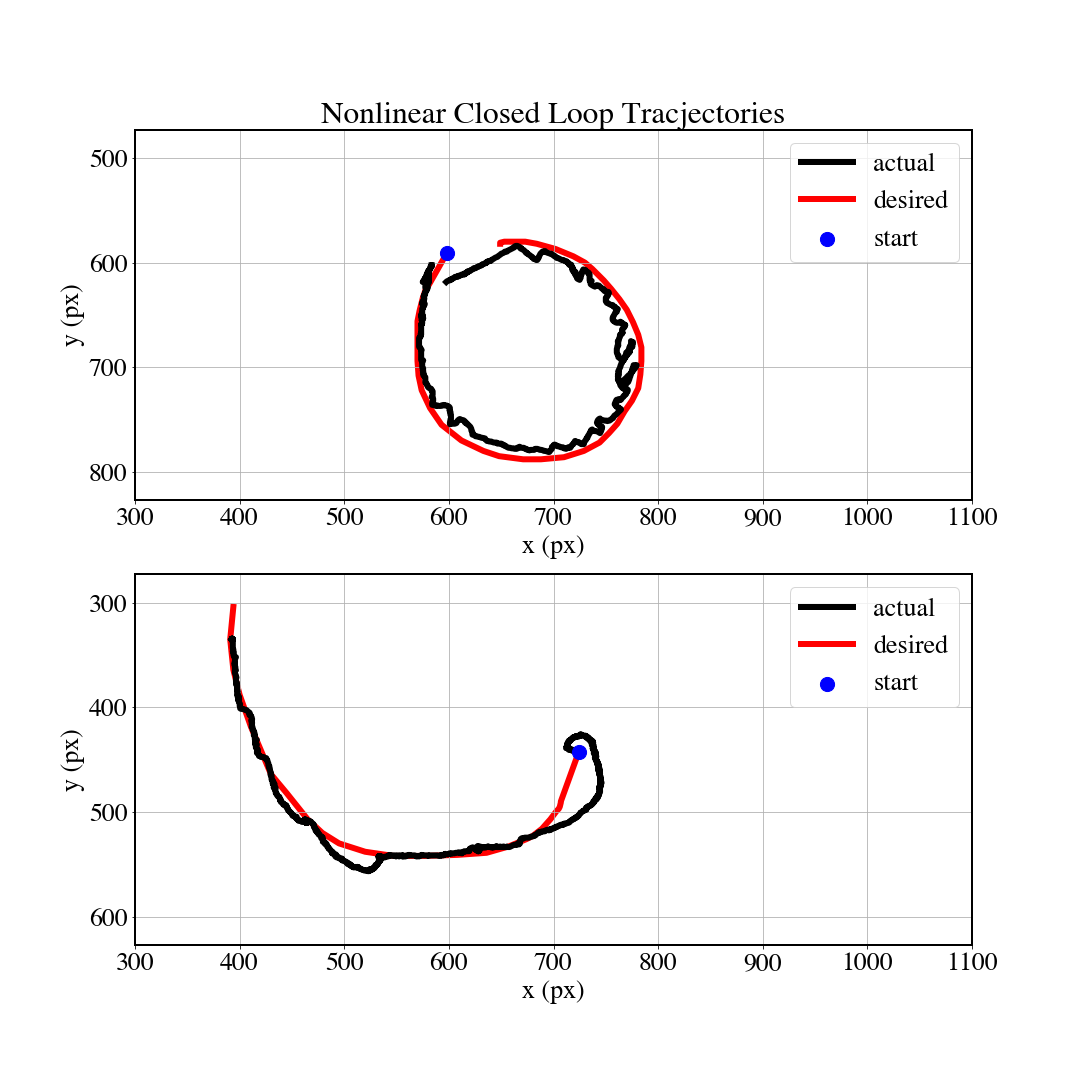}
  \caption {Two nonlinear trajectories implementing algorithm 1 and algorithm 2. The red curve indicates the desired trajectory defined by the users right mouse click and the black curve indicates the actual trajectory of the micro-motor.}
  \label{traj2}
\end{figure}

Using the above algorithm, a Janus micromotor can be directed to any target point. It is then straightforward to extend the algorithm to guide the microrobots along desired trajectories. Algorithm 2 allows a microrobot to follow predefined trajectories which the user can draw on the video display using the cursor (see figure \ref{traj2} and video 1). The trajectory is broken down into multiple nodes and the microrobot is guided toward each node in succession. Algorithm 2 accepts an array of points describing the desired trajectory. The microrobot's position is then found relative to the target position using custom tracking software in python. The error between the robots current position and target node position is determined. If the error is less than a user defined threshold value, the robot is considered to have arrived at the current node, and can move on to the next node. If the error is not less than this threshold, then the methods of algorithm 1 are used to guide the microrobot toward the target node.

Figure \ref{traj} shows four different micro-robotic trajectories. The blue dot represents the starting position of the microrobot. Algorithm 1 is applied to direct the microrobot towards the each node along the red line. The microrobots do not follow a direct path towards the target node, possibly due to changes in the magnetic moment of the paramagnetic microrobots caused by the changing magnetic field. Once the microrobots reach the target location, the code successfully keeps it in close proximity as its self-propulsion keeps the microrobot in continuous motion.

\section{Conclusion}
In summary, we have demonstrated a simple closed-loop control scheme for chemically powered Janus microrobots. We showed that using our magnetic setup and algorithms, catalytically propelled Janus colloids with an initially unknown magnetic moment orientation can be guided to a user defined target location. We expect that this work will be useful for the applications of Janus micromotors in situations where precise control in particle motion is required. We demonstrated that the microrobots follow more general curved trajectories that were drawn by the user. The closed-loop system could also be used to guide microrobots along trajectories that are determined by path-planning, such as for obstacle avoidance.

% \section*{Acknowledgment}

% The authors gratefully acknowledge the late Richard West for his help with the cell lines. This project was supported by the Delaware INBRE program, with a grant from the National Institute of General Medical Sciences – NIGMS (P20 GM103446) from the National Institutes of Health and the State of Delaware. This work was also supported by NSF grant OIA2020973. This content is solely the responsibility of the authors and does not necessarily represent the official views of NIH.

\end{document}